\documentclass[letterpaper]{article} 
\usepackage[preprint]{aaai2027}
\usepackage[hyphens]{url} 
\usepackage{graphicx} 
\usepackage{natbib} 
\usepackage{caption} 
\usepackage{amsmath}
\usepackage{amssymb}
\usepackage{booktabs}
\usepackage{multirow}
\usepackage{placeins}

\title{When Guidance Goes Off-Scale: Recalibrating Diffusion Transformers under Analog Compute-in-Memory Nonidealities}
\author{Wenshuai Yao\equalcontrib\textsuperscript{1},
Wenyong Zhou\equalcontrib\textsuperscript{2}}
\affiliations{\textsuperscript{1}School of Integrated Circuits, Peking University, Beijing, China\\
\textsuperscript{2}Department of Electrical and Computer Engineering, The University of Hong Kong, Hong Kong SAR, China}

\begin{document}

\maketitle

\begin{abstract}
Diffusion Transformers (DiTs) incur high memory traffic and energy costs because sampling repeatedly evaluates large denoisers dominated by linear operations. Analog compute-in-memory (CIM) can alleviate these costs by executing linear operations within weight-storing memory arrays. However, CIM nonidealities perturb effective weights, with errors accumulating along the state-dependent denoising trajectory; their interaction with classifier-free guidance (CFG) remains underexplored.
In this paper, we characterize the impact of analog CIM nonidealities on DiT sampling. Although conditional and unconditional predictions can each remain close to their clean counterparts, their difference---the CFG residual---is disproportionately attenuated and rotated. Identifying this residual as a controllable failure channel, we propose a retraining-free, sampler-side recalibration that adjusts only the CFG scale for a given CIM condition. Trajectory-level analysis shows that moderate recalibration strengthens the target-oriented component preserved in the distorted residual, enabling earlier commitment to a prompt-consistent semantic region. In contrast, excessive guidance amplifies the full noisy residual and degrades quality, resulting in a finite, noise-dependent optimum.
Extensive experiments on PixArt-$\Sigma$, PixArt-$\alpha$, and DiT-XL/2 show that the optimal guidance scale increases with CIM noise. Using 30,000 samples per condition, guidance recalibration consistently restores generation quality across simulated CIM mappings, closing at least $87\%$ of the CIM-induced FID gap at $\sigma_{\mathrm{CIM}}=0.20$. It reduces FID from $59.22$ to $20.49$ on PixArt-$\Sigma$, $72.37$ to $21.12$ on PixArt-$\alpha$, and $20.89$ to $6.62$ on DiT-XL/2.
\end{abstract}

\section{Introduction}

Diffusion models have become a leading paradigm for high-fidelity image generation \cite{ho2020ddpm}, and Diffusion Transformers (DiTs) demonstrate that transformer backbones can scale this paradigm effectively for both class-conditional and text-to-image synthesis \cite{peebles2023dit, chen2024pixartalpha, chen2024pixartsigma}. Achieving this quality, however, incurs substantial hardware cost. DiT sampling requires repeated evaluations of a large denoiser over many timesteps and repeated movement of large weight matrices between memory and compute units for attention projections and feed-forward layers. Analog compute-in-memory (CIM) is therefore an attractive deployment substrate: by performing linear operations within memory arrays, it can reduce costly data movement and improve energy efficiency \cite{shafiee2016isaac, sebastian2020imc}.

\begin{figure}[!t]
\centering
\includegraphics[width=\columnwidth]{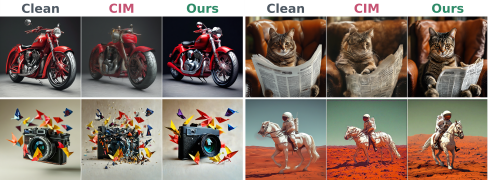}
\caption{Qualitative comparison across four prompts. Each triplet shows clean generation, CIM with clean-selected guidance, and CIM with recalibrated guidance.}
\label{fig:teaser}
\end{figure}

Analog CIM computation is subject to numerical nonidealities. Programming variation, conductance drift, read variation, and mixed-signal peripheral effects perturb the effective weights and computations realized by the deployed network \cite{sebastian2020imc, joshi2020pcm}. Such perturbations are particularly consequential for diffusion sampling. Unlike a feed-forward prediction task, a denoiser is evaluated repeatedly in a closed loop \cite{song2021score}. An error at one timestep therefore changes the latent state supplied to all subsequent timesteps, causing hardware-induced deviations to accumulate along a state-dependent denoising trajectory. While recent work has explored CIM acceleration for diffusion workloads \cite{jing2024aigcim}, the generation-quality degradation of modern pretrained DiTs under analog CIM nonidealities remains insufficiently understood. In particular, it is unclear how these nonidealities affect classifier-free guidance (CFG) \cite{ho2022cfg}, the primary conditional control mechanism used in diffusion sampling.

CFG combines unconditional and conditional denoiser predictions as
\begin{equation}
    \epsilon_w(x_t,t)
    =
    \epsilon_u(x_t,t)
    +
    w\bigl(
        \epsilon_c(x_t,t)
        -
        \epsilon_u(x_t,t)
    \bigr),
    \label{eq:cfg_intro}
\end{equation}
where $w$ is the guidance scale and
$g_t=\epsilon_c(x_t,t)-\epsilon_u(x_t,t)$
is the CFG residual. Reusing a guidance scale selected under clean digital inference after deployment implicitly assumes that the CFG residual remains sufficiently stable for the clean guidance calibration to transfer to CIM inference. Our analysis shows that this assumption can fail under CIM nonidealities. Although the conditional and unconditional predictions can each remain relatively close to their clean counterparts, their smaller difference---the CFG residual---is disproportionately distorted in relative magnitude and direction. This subtraction-sensitive distortion makes the clean-selected guidance scale miscalibrated under analog CIM nonidealities.

This observation identifies the CFG residual as a failure channel and the guidance scale as an available sampler-side control variable. We therefore propose \emph{guidance recalibration}, a retraining-free, sampler-side procedure that selects only the CFG scale for a target CIM operating condition. The pretrained denoiser, CIM mapping, scheduler, and sampling budget all remain unchanged. Figure~\ref{fig:teaser} illustrates the resulting recovery: compared with clean generation, CIM inference using the clean-selected scale exhibits substantial degradation, whereas recalibrating the existing CFG scale restores image structure and prompt-consistent details.

We view CFG as one-dimensional control of a state-dependent sampling trajectory. Under analog CIM nonidealities, the noisy CFG residual retains a useful component aligned with clean conditional control, but it also contains differential residual errors; meanwhile, common drift in the denoiser branches cannot be removed by changing the guidance scale alone. Moderate recalibration amplifies the retained useful component sufficiently to steer the trajectory toward the target semantic basin. Once the trajectory reaches its neighborhood, the learned denoising field provides local attraction toward that basin, and subsequent steps continue to refine the same semantic content. In contrast, excessive guidance amplifies both useful control and residual distortion, eventually driving the trajectory toward over-conditioned or low-quality regions. 
Thus, recalibration need not reproduce the clean trajectory at every timestep; rather, it restores a favorable trajectory-level operating point under the target CIM condition. 
Our contributions are threefold:
\begin{itemize}
\item We identify a subtraction-sensitive failure channel in CIM-deployed DiTs: the CFG residual is disproportionately distorted despite relatively stable conditional and unconditional predictions.

\item We propose a retraining-free, sampler-side guidance recalibration that adjusts only the CFG scale for a target CIM condition, without modifying the model or sampler.
    
\item We provide a trajectory-level account of the finite, noise-dependent guidance optimum and support it through residual, timestep, intervention, layer, seed, and perturbation analyses across three DiT models.
\end{itemize}

\section{Related Work}

\paragraph{Diffusion Transformers.}
Diffusion models generate samples through iterative denoising \cite{ho2020ddpm}, and Diffusion Transformers (DiTs) use scalable transformer backbones as denoisers \cite{peebles2023dit}. Representative DiT models span both class-conditional generation and large-scale text-to-image synthesis, including DiT-XL/2, PixArt-$\alpha$, and PixArt-$\Sigma$ \cite{peebles2023dit,chen2024pixartalpha,chen2024pixartsigma}. Their repeated attention and feed-forward projections make sampling computationally and memory intensive, motivating efficient hardware deployment. We study how analog CIM nonidealities affect this iterative DiT sampling process.

\paragraph{Analog CIM for Generative Models.}
Compute-in-memory (CIM) reduces data movement by executing matrix operations within memory arrays \cite{shafiee2016isaac,sebastian2020imc}. Prior work has explored CIM and heterogeneous acceleration for diffusion workloads, including diffusion-specific dataflows, mixed-precision designs, and analog implementations of score-based generation \cite{jing2024aigcim,zhu2025cim,guo2026denim,yang2026resistive}. For LLM deployment, NORA rescales linear operations to redistribute analog error among inputs, outputs, and weights \cite{hou2025nora}. However, analog CIM nonidealities, such as programming variation, drift, and read noise, perturb the effective deployed weights and computations \cite{joshi2020pcm,sebastian2020imc}. Existing studies primarily characterize hardware efficiency or aggregate generation degradation. In contrast, we identify how persistent CIM-induced perturbations distort the CFG residual control channel in modern pretrained DiTs and show that this diffusion-specific distortion can be mitigated through sampler-side guidance recalibration.

\paragraph{Classifier-Free Guidance.}
Classifier-free guidance (CFG) combines conditional and unconditional predictions to improve condition adherence during diffusion sampling \cite{ho2022cfg}. Several methods refine CFG under clean digital inference by controlling prediction magnitude, residual geometry, or the timesteps at which guidance is applied, including CFG Rescale, APG, Limited-Interval Guidance, C\({}^2\)FG, and CFG++ \cite{lin2024cfgrescale,sadat2025apg,kynkaanniemi2024interval,gao2026c2fg,chung2025cfgpp}. Recent analysis also interprets CFG through predictor-corrector dynamics \cite{bradley2025predictor}. These methods improve the trade-off between quality and alignment or reduce artifacts and off-manifold behavior under standard digital inference. Our work instead considers a deployment-induced miscalibration problem: analog CIM nonidealities distort the CFG residual itself, making the clean-selected guidance scale suboptimal. Rather than introducing a new guidance rule, we recalibrate the existing scalar guidance scale for the target CIM operating condition.

\section{Guidance Recalibration under Analog CIM}
\label{sec:method}

We first formulate hybrid analog CIM deployment and CFG sampling. We then identify the CFG residual as a subtraction-sensitive failure channel under CIM nonidealities, recalibrate its scalar control gain, and provide a trajectory-level interpretation of the resulting finite guidance optimum based on semantic basin dynamics.

\subsection{CIM Deployment and CFG}
\label{sec:cim_cfg}

\begin{figure}[!t]
\centering
\IfFileExists{figures/fig2_cim_deployment.pdf}{%
    \includegraphics[width=\columnwidth]{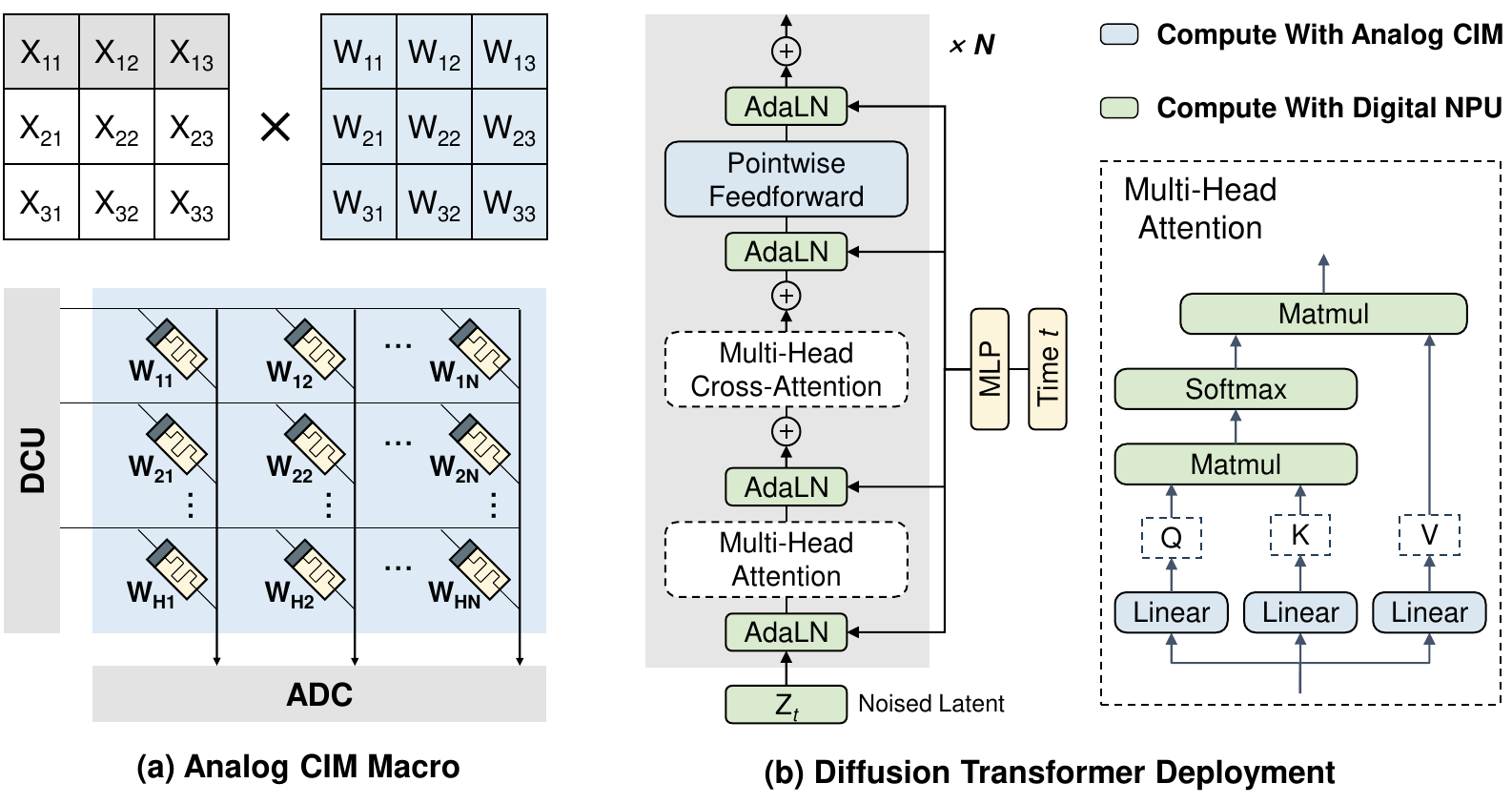}%
}{%
    \fbox{%
        \parbox[c][0.12\textheight][c]{0.92\columnwidth}{%
            \centering
            \textbf{CIM--DiT Deployment Schematic}\\[3pt]
            Placeholder for CIM mapping of attention and feed-forward
            linear layers.
        }%
    }%
}
\caption{Hybrid analog CIM deployment. (a) CIM-based matrix multiplication. (b) DiT mapping: linear layers are executed on CIM arrays, while other operations remain digital.}
\label{fig:cim_deployment}
\vspace{-0.2cm}
\end{figure}

We consider a hybrid analog CIM deployment of DiTs, illustrated in Figure~\ref{fig:cim_deployment}. Static linear weights in attention projections and feed-forward layers are mapped to CIM arrays, whereas normalization, nonlinear activations, softmax, data-dependent attention products, and scheduler updates remain digital. This setting preserves the original DiT architecture while accelerating the matrix operations that dominate its repeated denoiser evaluations.

Motivated by hardware studies that model conductance variability through Gaussian weight perturbations \cite{joshi2020pcm,wan2022neurram}, we perturb each mapped linear weight matrix $W_\ell$ as
\begin{equation}
\widetilde{W}_{\ell} = W_{\ell}\odot\left(1+\sigma_{\mathrm{CIM}}\Xi_{\ell}\right), \qquad [\Xi_\ell]_{ij}\overset{\mathrm{i.i.d.}}{\sim}\mathcal{N}(0,1),
\label{eq:cim_weight_noise}
\end{equation}
where $\sigma_{\mathrm{CIM}}$ controls the perturbation severity. A sampled realization is fixed throughout a denoising trajectory, reflecting persistent weight-related deviations in a deployed CIM array. Equation~\eqref{eq:cim_weight_noise} is an algorithm-level proxy for analog CIM nonidealities rather than a complete circuit-level model.

Classifier-free guidance (CFG) combines unconditional and conditional denoiser predictions as
\begin{equation}
\epsilon_{w,t} = u_t + w g_t, \qquad g_t = c_t-u_t,
\label{eq:cfg}
\end{equation}
where $u_t=\epsilon_\theta(x_t,t,\emptyset)$ and $c_t=\epsilon_\theta(x_t,t,y)$ denote the unconditional and conditional predictions, respectively. The scalar $w$ controls the strength of the conditional residual $g_t$. Let $w_0$ denote the scale selected under clean digital inference. A straightforward deployment baseline reuses $w_0$ after mapping the model to CIM; this baseline can become miscalibrated under CIM nonidealities.

\subsection{CFG Residual Distortion}
\label{sec:residual_distortion}

Let $\widetilde{\theta}(\xi)$ denote the effective weights under a CIM realization $\xi$. At a given latent state $x_t$, the induced prediction perturbation is
\begin{equation}
\delta_t(x_t,y;\xi) = \epsilon_{\widetilde{\theta}(\xi)}(x_t,t,y)-\epsilon_{\theta}(x_t,t,y).
\label{eq:prediction_perturbation}
\end{equation}
Although Equation~\eqref{eq:cim_weight_noise} is Gaussian at the weight level, the resulting output perturbation generally depends on the state, timestep, condition, and CIM realization.

For the two CFG branches, we write
\begin{equation}
\widetilde{u}_t=u_t+\delta_{u,t}, \qquad \widetilde{c}_t=c_t+\delta_{c,t}.
\label{eq:branch_perturbation}
\end{equation}
Their noisy residual is therefore
\begin{equation}
\widetilde{g}_t=\widetilde{c}_t-\widetilde{u}_t=g_t+\delta_{g,t}, \qquad \delta_{g,t}=\delta_{c,t}-\delta_{u,t}.
\label{eq:residual_perturbation}
\end{equation}
This subtraction exposes differential branch errors. Even when $\delta_{u,t}$ and $\delta_{c,t}$ are small relative to the branch predictions, their difference can be substantial relative to the smaller residual $g_t$.
\begin{figure}[!t]
\centering
\includegraphics[width=\columnwidth]{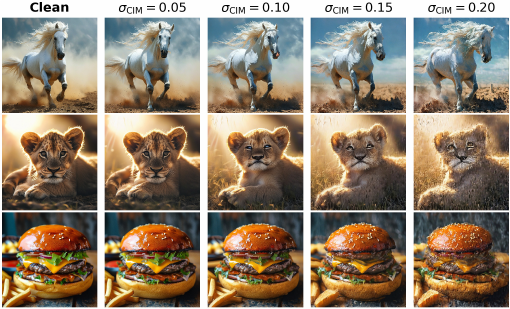}
\caption{PixArt-$\Sigma$ generations at $1024\times1024$ resolution under increasing CIM nonidealities at the clean-selected scale $w_0=2.0$.}
\label{fig:cim-noise-drift}
\vspace{-0.2cm}
\end{figure}
\begin{figure}[!t]
\centering
\includegraphics[width=\columnwidth]{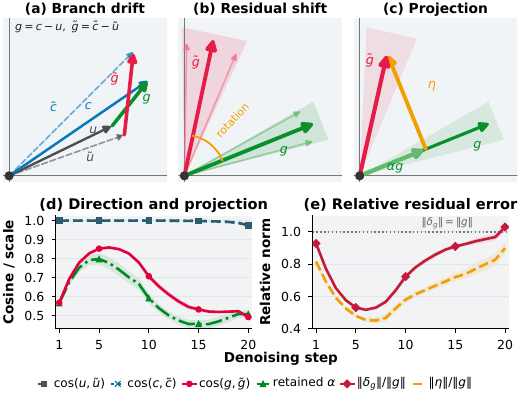}
\caption{CFG and residual diagnostics under CIM nonidealities. (a) Branch perturbations. (b) Residual rotation. (c) Clean-aligned decomposition. (d--e) Timestep statistics at $\sigma_{\mathrm{CIM}}=0.20$; shading indicates 95\% confidence intervals.}
\label{fig:residual-mechanism}
\vspace{-0.2cm}
\end{figure}

Figure~\ref{fig:cim-noise-drift} visualizes generation at the clean-selected guidance scale as the CIM perturbation level increases.
\begin{figure*}[!t]
\centering
\includegraphics[width=\textwidth]{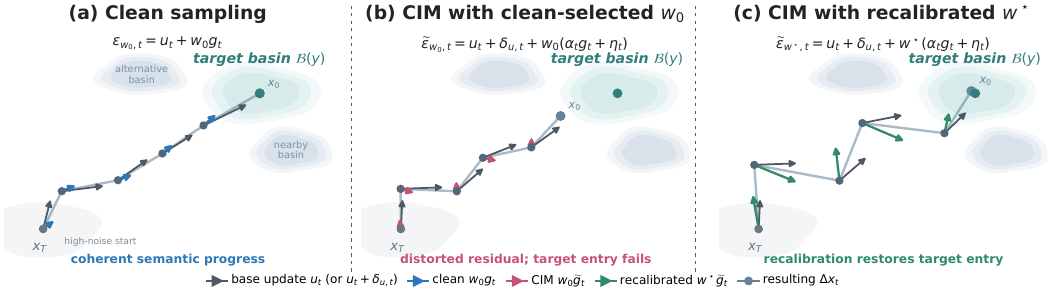}
\caption{Conceptual sampling trajectories under (a) clean inference, (b) CIM inference with the clean-selected scale $w_0$, and (c) CIM inference with the recalibrated scale $w^\star$. Shaded contours indicate semantic basins; arrows denote base and guidance updates. Each trajectory segment is the vector sum of the two component arrows.}
\label{fig:semantic-basin}
\vspace{-0.4cm}
\end{figure*}

We characterize this distortion by decomposing the noisy residual into a component aligned with the clean residual and an orthogonal component:
\begin{equation}
\widetilde{g}_t=\alpha_t g_t+\eta_t, \qquad \alpha_t=\frac{\langle \widetilde{g}_t,g_t\rangle}{\lVert g_t\rVert_2^2}, \qquad \langle \eta_t,g_t\rangle=0.
\label{eq:residual_decomposition}
\end{equation}
Here, $\alpha_t g_t$ is the retained clean-aligned component, while $\eta_t$ denotes the residual component orthogonal to $g_t$. Both terms vary over timesteps, latent states, prompts, and CIM realizations.

Figure~\ref{fig:residual-mechanism} shows that the conditional and unconditional branches remain substantially closer to their clean counterparts than does their difference. In particular, the noisy residual exhibits reduced directional agreement, attenuated aligned magnitude, and large relative error. Thus, beyond introducing generic denoiser drift, CIM nonidealities disproportionately distort the control channel directly scaled by CFG.

Substituting Equation~\eqref{eq:residual_decomposition} into CFG gives
\begin{equation}
\widetilde{\epsilon}_{w,t}=u_t+\delta_{u,t}+w\left(\alpha_tg_t+\eta_t\right).
\label{eq:noisy_cfg}
\end{equation}
Equation~\eqref{eq:noisy_cfg} identifies a controllable trade-off: changing $w$ cannot directly remove the common branch drift $\delta_{u,t}$, but it simultaneously scales the retained useful component $\alpha_tg_t$ and the orthogonal residual component $\eta_t$.

\subsection{Guidance Recalibration}
\label{sec:guidance_recalibration}

Motivated by this diagnosis, we recalibrate only the existing CFG scale for a target CIM operating condition. The pretrained denoiser, CIM mapping, scheduler, and sampling budget remain fixed. Given an independent calibration set $\mathcal{D}_{\mathrm{cal}}$ and a candidate set of guidance scales $\mathcal{W}$, we select
\begin{equation}
w^\star(\mathcal{C})=\arg\min_{w\in\mathcal{W}}\mathcal{L}\bigl(w;\mathcal{D}_{\mathrm{cal}},\mathcal{C}\bigr),
\label{eq:guidance_recalibration}
\end{equation}
where $\mathcal{C}$ denotes the target CIM operating condition, including the mapped layers, nonideality model, and perturbation level. In our main experiments, $\mathcal{L}$ is FID and $\mathcal{C}$ is indexed by $\sigma_{\mathrm{CIM}}$; we write the corresponding selected scale as $w_\sigma^\star$.

The selected scale $w^\star$ is fixed for all subsequent samples under the same operating condition. This procedure requires no retraining, weight update, activation correction, per-prompt adaptation, or additional denoiser evaluation during sampling. The sweep is performed once per target operating condition; after selection, deployment retains the original sampling procedure and uses $w^\star$ as a fixed scalar setting. Rather than attempting to invert CIM errors or reproduce the clean trajectory at individual timesteps, guidance recalibration selects a favorable operating point for the noisy closed-loop sampler.

\subsection{Trajectory Interpretation}
\label{sec:trajectory_interpretation}

Guidance recalibration acts on a closed-loop sampling process. For a generic reverse sampler,
\begin{equation}
\widetilde{x}^{(w)}_{t-1}=S_t\left(\widetilde{x}^{(w)}_t,\widetilde{\epsilon}_{w,t}\right),
\label{eq:noisy_sampler}
\end{equation}
changing $w$ affects not only the current update but also the latent states supplied to all subsequent denoiser evaluations. Recalibration is therefore a trajectory-level control problem rather than pointwise matching of clean denoiser predictions.

Defining $\Delta x_t^{(w)}=\widetilde{x}_{t-1}^{(w)}-\widetilde{x}_t^{(w)}$ and locally linearizing the sampler around the noisy unconditional prediction at the current state gives
\begin{equation}
\Delta x^{(w)}_t\approx v_{u,t}+w r_t, \qquad r_t=B_t\widetilde{g}_t,
\label{eq:local_control}
\end{equation}
where $v_{u,t}$ is the base update, $B_t$ is the local prediction-to-state mapping induced by the sampler, and $r_t$ is the sampler-propagated residual control. When $r_t$ retains a component that promotes target-consistent structure over semantically informative steps, increasing $w$ can strengthen this retained control and help the trajectory reach a region in which the target semantics become locally persistent.

Figure~\ref{fig:semantic-basin} provides a conceptual interpretation. Under CIM nonidealities, the clean-selected scale $w_0$ can yield insufficient target-oriented control because the residual is attenuated and rotated. A moderately larger scale $w^\star$ amplifies the useful component that remains, allowing the trajectory to reach a target-consistent region earlier. Once target-consistent structure becomes locally dominant, the learned denoising dynamics tend to preserve and refine that structure over subsequent steps, yielding attractor-like local behavior. Related attractor-like behavior has been used to characterize persistent diffusion trajectories in memorization settings \cite{jain2025attraction}; here, \emph{semantic basin} denotes the local persistence of target-consistent structure. This account predicts that recalibration should advance persistent semantic entry, which we examine through the timestep analysis in Figure~\ref{fig:timestep-semantic-entry}.

\begin{table*}[t]
\centering
{%
\small
\setlength{\tabcolsep}{2.15pt}
\renewcommand{\arraystretch}{1.10}
\begin{tabular}{@{}l@{\hspace{2.0pt}}cccccccc@{}}
\toprule
Model and benchmark & \(\sigma_{\mathrm{CIM}}\) & CFG \(w\) & FID\(\downarrow\) & KID\(\times10^3\downarrow\) & Alignment\(\uparrow\) & Precision\(\uparrow\) & Density\(\uparrow\) & Coverage\(\uparrow\) \\
\midrule
\multirow{5}{*}{\shortstack[l]{PixArt-\(\Sigma\) (\(512^2\), COCO)}} & 0 (clean) & 1.5 & 20.51 & 7.75 & 30.34 & 0.550 & 0.654 & 0.595 \\
 & 0.05 & 1.5 (\(=w_0\)) & 20.31 & 7.68 & 30.31 & 0.551 & 0.647 & 0.582 \\
 & 0.10 & 1.5 \(\!\rightarrow\!\) 2.0 & 21.12 \(\!\rightarrow\!\) \underline{\textbf{19.98}} & 8.27 \(\!\rightarrow\!\) \underline{\textbf{7.05}} & 30.09 \(\!\rightarrow\!\) \underline{\textbf{30.69}} & 0.513 \(\!\rightarrow\!\) \underline{\textbf{0.552}} & 0.575 \(\!\rightarrow\!\) \underline{\textbf{0.654}} & 0.556 \(\!\rightarrow\!\) \underline{\textbf{0.580}} \\
 & 0.15 & 1.5 \(\!\rightarrow\!\) 2.5 & 28.62 \(\!\rightarrow\!\) \underline{\textbf{20.18}} & 12.85 \(\!\rightarrow\!\) \underline{\textbf{6.87}} & 29.37 \(\!\rightarrow\!\) \underline{\textbf{30.79}} & 0.432 \(\!\rightarrow\!\) \underline{\textbf{0.543}} & 0.423 \(\!\rightarrow\!\) \underline{\textbf{0.628}} & 0.477 \(\!\rightarrow\!\) \underline{\textbf{0.572}} \\
 & 0.20 & 1.5 \(\!\rightarrow\!\) 4.5 & 59.22 \(\!\rightarrow\!\) \underline{\textbf{20.49}} & 30.41 \(\!\rightarrow\!\) \underline{\textbf{6.49}} & 27.32 \(\!\rightarrow\!\) \underline{\textbf{31.17}} & 0.248 \(\!\rightarrow\!\) \underline{\textbf{0.527}} & 0.203 \(\!\rightarrow\!\) \underline{\textbf{0.602}} & 0.292 \(\!\rightarrow\!\) \underline{\textbf{0.560}} \\
\midrule
\multirow{5}{*}{\shortstack[l]{PixArt-\(\alpha\) (\(512^2\), COCO)}} & 0 (clean) & 1.5 & 21.63 & 8.21 & 29.98 & 0.545 & 0.624 & 0.567 \\
 & 0.05 & 1.5 \(\!\rightarrow\!\) 2.0 & 21.84 \(\!\rightarrow\!\) \underline{\textbf{21.01}} & 8.50 \(\!\rightarrow\!\) \underline{\textbf{7.32}} & 29.89 \(\!\rightarrow\!\) \underline{\textbf{30.48}} & 0.533 \(\!\rightarrow\!\) \underline{\textbf{0.571}} & 0.616 \(\!\rightarrow\!\) \underline{\textbf{0.717}} & 0.561 \(\!\rightarrow\!\) \underline{\textbf{0.587}} \\
 & 0.10 & 1.5 \(\!\rightarrow\!\) 2.5 & 24.16 \(\!\rightarrow\!\) \underline{\textbf{20.75}} & 10.34 \(\!\rightarrow\!\) \underline{\textbf{7.10}} & 29.52 \(\!\rightarrow\!\) \underline{\textbf{30.65}} & 0.495 \(\!\rightarrow\!\) \underline{\textbf{0.576}} & 0.537 \(\!\rightarrow\!\) \underline{\textbf{0.716}} & 0.514 \(\!\rightarrow\!\) \underline{\textbf{0.589}} \\
 & 0.15 & 1.5 \(\!\rightarrow\!\) 3.0 & 35.44 \(\!\rightarrow\!\) \underline{\textbf{20.71}} & 17.68 \(\!\rightarrow\!\) \underline{\textbf{7.10}} & 28.53 \(\!\rightarrow\!\) \underline{\textbf{30.60}} & 0.385 \(\!\rightarrow\!\) \underline{\textbf{0.558}} & 0.368 \(\!\rightarrow\!\) \underline{\textbf{0.655}} & 0.418 \(\!\rightarrow\!\) \underline{\textbf{0.566}} \\
 & 0.20 & 1.5 \(\!\rightarrow\!\) 5.0 & 72.37 \(\!\rightarrow\!\) \underline{\textbf{21.12}} & 40.60 \(\!\rightarrow\!\) \underline{\textbf{7.39}} & 25.95 \(\!\rightarrow\!\) \underline{\textbf{30.68}} & 0.212 \(\!\rightarrow\!\) \underline{\textbf{0.517}} & 0.148 \(\!\rightarrow\!\) \underline{\textbf{0.609}} & 0.208 \(\!\rightarrow\!\) \underline{\textbf{0.536}} \\
\midrule
\multirow{5}{*}{\shortstack[l]{DiT-XL/2 (\(256^2\), ImageNet)}} & 0 (clean) & 1.3 & 4.49 & 0.85 & 82.77 & 0.822 & 1.060 & 0.872 \\
 & 0.05 & 1.3 (\(=w_0\)) & 4.51 & 0.82 & 81.60 & 0.807 & 1.039 & 0.857 \\
 & 0.10 & 1.3 \(\!\rightarrow\!\) 1.4 & 5.13 \(\!\rightarrow\!\) \underline{\textbf{4.66}} & 1.03 \(\!\rightarrow\!\) \underline{\textbf{0.83}} & 79.10 \(\!\rightarrow\!\) \underline{\textbf{82.40}} & 0.776 \(\!\rightarrow\!\) \underline{\textbf{0.805}} & 0.946 \(\!\rightarrow\!\) \underline{\textbf{0.995}} & 0.828 \(\!\rightarrow\!\) \underline{\textbf{0.845}} \\
 & 0.15 & 1.3 \(\!\rightarrow\!\) 1.7 & 8.65 \(\!\rightarrow\!\) \underline{\textbf{5.20}} & 3.24 \(\!\rightarrow\!\) \underline{\textbf{1.24}} & 72.63 \(\!\rightarrow\!\) \underline{\textbf{85.76}} & 0.670 \(\!\rightarrow\!\) \underline{\textbf{0.797}} & 0.790 \(\!\rightarrow\!\) \underline{\textbf{0.991}} & 0.768 \(\!\rightarrow\!\) \underline{\textbf{0.855}} \\
 & 0.20 & 1.3 \(\!\rightarrow\!\) 2.0 & 20.89 \(\!\rightarrow\!\) \underline{\textbf{6.62}} & 11.68 \(\!\rightarrow\!\) \underline{\textbf{1.39}} & 59.28 \(\!\rightarrow\!\) \underline{\textbf{84.91}} & 0.537 \(\!\rightarrow\!\) \underline{\textbf{0.756}} & 0.563 \(\!\rightarrow\!\) \underline{\textbf{0.880}} & 0.645 \(\!\rightarrow\!\) \underline{\textbf{0.819}} \\
\bottomrule
\end{tabular}
}
\caption{Large-scale population-level evaluation across three diffusion transformers and multiple simulated CIM mappings, with 30,000 samples per setting. Noisy entries report \(w_0\!\rightarrow w_\sigma^\star\); a single value is shown when the selected scales coincide. Underlining marks the recalibrated endpoint, and bold marks improvement over the clean-guidance baseline.}
\label{tab:main-30k}
\end{table*}

The benefit is bounded: excessive guidance amplifies the entire distorted residual, including components misaligned with target progress, while leaving the common drift $\delta_{u,t}$ uncontrolled. These errors accumulate along the closed-loop trajectory, causing over-conditioning and degraded samples. Thus, the optimal guidance scale is finite and depends on the CIM condition. This analysis predicts that the preferred scale shifts with CIM severity, recalibration accelerates semantic commitment, and both residual distortion and base-path drift contribute to degradation.

\section{Experiments}
\label{sec:experiments}

\subsection{Experimental Setup}

\begin{figure*}[!t]
\centering
\includegraphics[width=\textwidth]{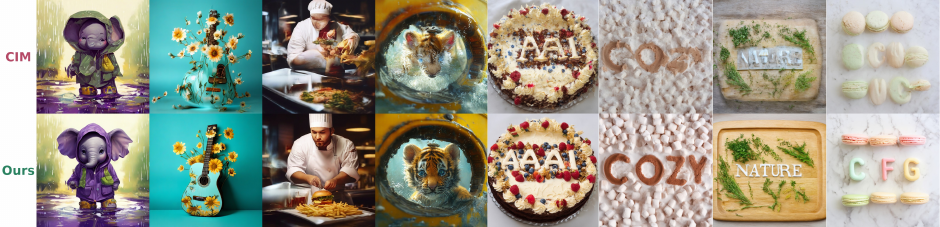}
\caption{Qualitative comparisons under CIM noise. Left: PixArt-\(\Sigma\) text-to-image generation at \(\sigma_{\mathrm{CIM}}=0.15\), using \(w_0=1.5\) and \(w_\sigma^\star=2.5\). Right: SD3.5 Medium text rendering at \(\sigma_{\mathrm{CIM}}=0.10\), using \(w_0=1.5\) and \(w_\sigma^\star=3.0\).}
\label{fig:qualitative-comparison}
\end{figure*}

\paragraph{Models and benchmarks.}
We evaluate PixArt-\(\Sigma\)-XL/2 and PixArt-\(\alpha\)-XL/2 at \(512^2\) for text-to-image generation and DiT-XL/2 at \(256^2\) for class-conditional generation. PixArt-\(\Sigma\) and Stable Diffusion 3.5 Medium (SD3.5 Medium) \cite{esser2024sd3,stabilityai2024sd35} at \(1024^2\) are used only for qualitative examples in Figures~\ref{fig:cim-noise-drift} and~\ref{fig:qualitative-comparison}. For standard evaluation, our 30,000-sample text-to-image setting uses MS-COCO 2014 validation captions and references \cite{lin2014coco}, while the 30,000-sample canonical class-conditional DiT setting uses ImageNet validation statistics \cite{deng2009imagenet}. Quantitative PixArt experiments use the default DPM-Solver scheduler \cite{lu2022dpmsolver} with 20 denoising steps, while DiT-XL/2 uses 50 steps. We first identify \(w_0\) under clean inference and \(w_\sigma^\star\) in an independent 3,000-sample guidance sweep, then hold both scales fixed for all subsequent 30,000-sample evaluations.

\paragraph{CIM protocol.}
Unless specified otherwise, Equation~\eqref{eq:cim_weight_noise} is applied to the query, key, value, and output projections of self- and cross-attention and to both feed-forward linear layers. Each deterministic weight-noise draw represents an independently programmed virtual CIM device and remains fixed for all samples and denoising steps within its inference batch. Different batches are assigned to different virtual devices, yielding a population-level evaluation across CIM mappings. A single guidance scale is shared across the entire device population without per-device adaptation. Clean-selected and recalibrated settings use identical prompts, sample seeds, CIM realizations, samplers, and mapped layer sets.

\paragraph{Metrics.}
We report FID \cite{heusel2017ttur} and KID \cite{binkowski2018demystifying} for distributional quality. Conditional alignment is CLIPScore \cite{radford2021clip,hessel2021clipscore} for text-to-image models and ImageNet ViT-B/16 \cite{dosovitskiy2021vit} top-1 accuracy for DiT-XL/2. Precision measures sample fidelity, while density and coverage characterize local realism and support coverage \cite{kynkaanniemi2019precision,naeem2020reliable}. DINO-clean is the paired DINOv2 cosine similarity \cite{oquab2023dinov2} between a CIM output and the clean output generated with $w_0$ from the same condition and sample seed. Accordingly, FID and related distributional metrics characterize the pooled output distribution of the simulated device population.

\begin{figure}[!t]
\centering
\includegraphics[width=\columnwidth]{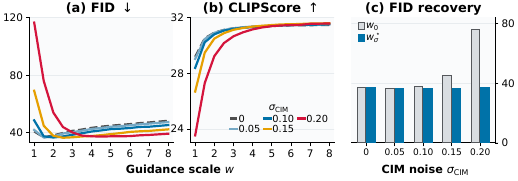}
\caption{Guidance response on a 3,000-prompt PixArt-\(\Sigma\) sweep. (a) FID and (b) CLIPScore versus $w$. (c) FID at $w_0$ and $w_\sigma^\star$.}
\label{fig:guidance-response}
\end{figure}

\subsection{Experimental Results}
\paragraph{Noise-Dependent Guidance Response}

Figure~\ref{fig:guidance-response} characterizes the response to \(w\) on 3,000 COCO 2017 validation prompts. The FID curves exhibit a finite minimum rather than monotonic improvement. The FID-optimal scale moves from \(1.5\) at \(\sigma_{\mathrm{CIM}}\in\{0,0.05\}\) to \(2.0\), \(2.5\), and \(4.5\) at noise levels \(0.10\), \(0.15\), and \(0.20\), respectively. At \(\sigma_{\mathrm{CIM}}=0.20\), recalibration lowers FID from \(75.94\) to \(37.42\), whereas further increases in \(w\) degrade FID. CLIPScore increases and then saturates as \(w\) grows, indicating that semantic alignment alone is insufficient for selecting the operating point: excessive guidance can preserve prompt alignment while degrading distributional fidelity. The selected scales are then fixed for the independent 30,000-sample evaluation below.

\paragraph{Calibration Efficiency and Transfer}

To assess whether guidance recalibration requires a large calibration set, we select the guidance scale using 256 calibration prompts and evaluate it on 2,000 disjoint prompts. At \(\sigma_{\mathrm{CIM}}=0.15\), recalibration reduces FID from \(51.09\) to \(43.92\), and at \(\sigma_{\mathrm{CIM}}=0.20\), from \(79.94\) to \(44.03\), compared with a clean reference of \(43.58\). The selected scale is shared across all held-out prompts, without per-prompt search or adaptation. Thus, a small calibration subset is sufficient to identify a transferable operating point, making recalibration a lightweight deployment-time procedure.

\paragraph{Large-Scale Generation Quality}

Using scales selected on independent 3,000-sample sweeps, Table~\ref{tab:main-30k} evaluates the pooled population-level distribution with 30,000 samples per setting. At \(\sigma_{\mathrm{CIM}}=0.20\), recalibration reduces FID from \(59.22\) to \(20.49\) on PixArt-\(\Sigma\), from \(72.37\) to \(21.12\) on PixArt-\(\alpha\), and from \(20.89\) to \(6.62\) on DiT-XL/2. KID changes in the same direction, while conditional alignment, precision, density, and coverage improve for all three models. For each model at \(\sigma_{\mathrm{CIM}}=0.20\), a single recalibrated scale shared across the simulated device population closes at least \(87\%\) of the CIM-induced FID gap relative to clean inference without modifying the noisy weights or sampling budget.

\begin{figure*}[!t]
\centering
\includegraphics[width=\textwidth]{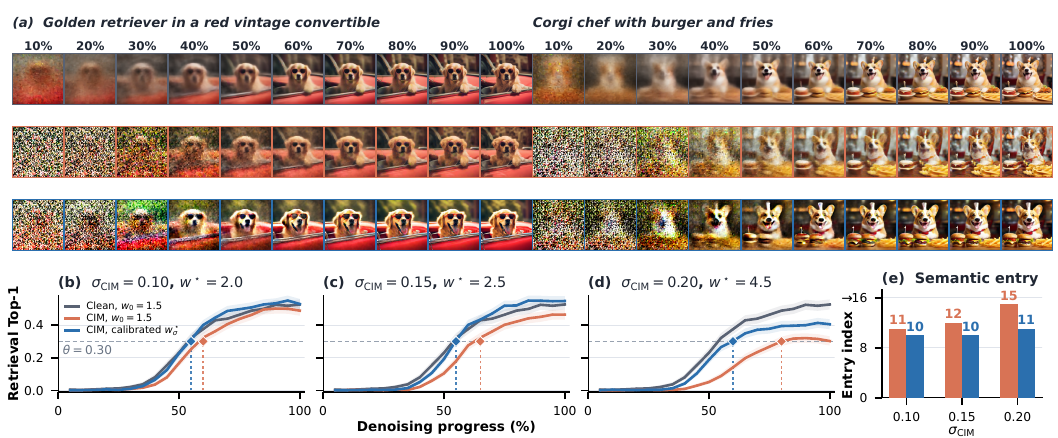}
\caption{Timestep semantic-entry analysis. (a) Ten evenly spaced checkpoints from 20-step predicted-clean trajectories for two prompts; rows show clean inference, CIM with \(w_0\), and CIM with \(w_\sigma^\star\). (b--d) Prompt retrieval top-1 accuracy across all steps. (e) Persistent semantic-entry indices at \(\theta=0.30\).}
\label{fig:timestep-semantic-entry}
\end{figure*}

\paragraph{Comparison with Sampler-Side Guidance Controls}

\begin{table}[t]
\centering
{%
\footnotesize
\setlength{\tabcolsep}{3.2pt}
\renewcommand{\arraystretch}{1.00}
\begin{tabular}{@{}lrr@{}}
\toprule
Method & \(\sigma_{\mathrm{CIM}}=0.15\) & \(\sigma_{\mathrm{CIM}}=0.20\) \\
\midrule
Clean-best CFG & 44.93 & 75.94 \\
CFG Rescale & 40.40 & 40.72 \\
APG & 39.21 & 41.08 \\
C\({}^2\)FG & 38.11 & 39.52 \\
Limited-Interval Guidance & 37.28 & 38.76 \\
\textbf{Fixed recalibration (ours)} & \textbf{36.60} & \textbf{37.61} \\
\bottomrule
\end{tabular}
}
\caption{FID in the shared 3,000-prompt PixArt-\(\Sigma\) comparison of sampler-side guidance controls.}
\label{tab:baseline-comparison}
\end{table}

Although not designed for CIM, these methods provide relevant sampler-side comparisons. We tune CFG Rescale \cite{lin2024cfgrescale}, APG \cite{sadat2025apg}, C\({}^2\)FG \cite{gao2026c2fg}, and Limited-Interval Guidance \cite{kynkaanniemi2024interval} under identical CIM settings and report the shared 3,000-prompt comparison. The clean-best row is shared with Figure~\ref{fig:guidance-response}, while fixed recalibration uses its preselected scale rather than the sweep optimum.

Table~\ref{tab:baseline-comparison} shows that all sampler-side controls improve over clean-best CFG, indicating that the guidance residual remains an effective correction channel under CIM noise. However, their correction forms were developed for generic guidance artifacts under standard inference rather than the noise-dependent residual distortion induced by CIM. Even after configuration search under each CIM condition, they remain less effective than directly recalibrating the residual strength for the target operating point. This ordering is consistent with guidance-scale mismatch being a substantial and directly correctable component of CIM-induced degradation.

\paragraph{Qualitative Comparisons}

Figure~\ref{fig:qualitative-comparison} provides paired qualitative comparisons. Recalibration improves object structure and prompt-specific details for general text-to-image prompts, while also producing more legible and coherent rendered text. These examples indicate that the recovery extends beyond aggregate distributional metrics to challenging semantic and structural details.

\begin{table*}[!t]
\centering
\begin{minipage}[t]{0.56\textwidth}
\vspace{0pt}
\centering
\small
\fontsize{8}{\baselineskip}\selectfont
\setlength{\tabcolsep}{2.0pt}
\renewcommand{\arraystretch}{1.00}
\begin{tabular*}{\linewidth}{@{\extracolsep{\fill}}lrrrrrrr@{}}
\toprule
\multicolumn{8}{c}{\textbf{(a) Endpoint counterfactuals}} \\
\cmidrule(lr){1-8}
Metric & Clean & CIM \(w_0\) & \textbf{CIM \(w^\star\)} & Restore \(u\) & Restore \(g\) &
Remove \(\eta\) & Restore \(\alpha\) \\
\midrule
FID\(\downarrow\) & 36.19 & 75.41 & \textbf{36.64} & 42.83 & 68.34 & 127.75 & 44.49 \\
CLIP\(\uparrow\) & 30.45 & 27.27 & \textbf{31.27} & 28.52 & 28.87 & 23.23 & 30.35 \\
DINO-clean\(\uparrow\) & 1.000 & 0.463 & 0.673 & 0.621 & 0.631 & 0.261 & \textbf{0.699} \\
\bottomrule
\end{tabular*}
\end{minipage}
\hfill
\begin{minipage}[t]{0.42\textwidth}
\vspace{0pt}
\centering
\small
\fontsize{8}{\baselineskip}\selectfont
\setlength{\tabcolsep}{2.2pt}
\renewcommand{\arraystretch}{1.00}
\begin{tabular*}{\linewidth}{@{\extracolsep{\fill}}lrrrrr@{}}
\toprule
\multicolumn{6}{c}{\textbf{(b) Continuous endpoint interpolation: FID\(\downarrow\)}} \\
\cmidrule(lr){1-6}
Intervention & \(0.00\) & \(0.25\) & \(0.50\) & \(0.75\) & \(1.00\) \\
\midrule
Restore aligned gain & 75.4 & 64.1 & 55.6 & 49.1 & \textbf{44.5} \\
Remove orthogonal term & \textbf{75.4} & 86.6 & 99.6 & 114.0 & 127.9 \\
Remove base error & 75.4 & 58.5 & 48.3 & 44.0 & \textbf{42.9} \\
\bottomrule
\end{tabular*}
\end{minipage}
\caption{Residual-surgery analysis at \(\sigma_{\mathrm{CIM}}=0.20\) over 3,000 prompts. (a) Endpoint counterfactuals. (b) FID along intervention paths from untreated CIM (\(0\)) to the full intervention (\(1\)).}
\label{tab:residual-surgery}
\end{table*}

\paragraph{Timestep Semantic Entry}

We probe whether quality recovery is accompanied by earlier semantic commitment during sampling. At each reverse step, we decode the scheduler-consistent predicted-clean sample \(\widehat{x}_{0\mid t}\). For each of 1,000 prompts, its CLIP image embedding retrieves the matching text from 1,000 candidates. We define the semantic entry index as the earliest reverse step, counted from the start of sampling, at which the aggregate top-1 retrieval rate reaches a threshold \(\theta\) and remains above it thereafter. Based on visual inspection of intermediate generations, we set \(\theta=0.30\), at which prompt-consistent content becomes persistently recognizable.

Figure~\ref{fig:timestep-semantic-entry}(a) visualizes ten evenly spaced checkpoints for two prompts under clean inference, CIM with \(w_0\), and CIM with \(w_\sigma^\star\). Panels (b--e) show that recalibration advances persistent entry from step \(11\) to \(10\) at \(\sigma_{\mathrm{CIM}}=0.10\), from \(12\) to \(10\) at \(0.15\), and from \(15\) to \(11\) at \(0.20\). The increasing gain with noise is consistent with the proposed trajectory-level interpretation: strengthening the retained target-oriented residual moves the sampler into a target-consistent semantic region earlier, after which the state-dependent denoising dynamics preserve and refine those semantics.

\paragraph{Counterfactual Residual Surgery}

To probe the decomposition \(\widetilde g=\alpha g+\eta\) through interventions, Table~\ref{tab:residual-surgery} applies counterfactual modifications to paired clean and noisy predictions at \(\sigma_{\mathrm{CIM}}=0.20\). The endpoint interventions restore the clean base prediction \(u\), the clean residual \(g\), or the clean-aligned gain \(\alpha\), and separately remove the orthogonal component \(\eta\). Continuous paths vary each intervention from untreated CIM at strength \(0\) to the full intervention at strength \(1\), reducing dependence on any single endpoint.

Restoring \(u\) lowers FID from \(75.41\) to \(42.83\), while restoring \(\alpha\) yields \(44.49\) and improves monotonically as the intervention strength increases. Likewise, progressively removing base-path error improves FID, indicating that both common drift and weakened target-oriented residual control contribute to degradation. Removing \(\eta\) alone worsens FID, indicating that its effect is coupled to base drift and trajectory feedback rather than independently removable. Recalibration reaches \(36.64\), close to the clean FID of \(36.19\), by scaling the intact noisy residual instead of attempting componentwise cancellation.

\section{Conclusion}
In this paper, we systematically studied the impact of analog CIM nonidealities on DiT sampling and identified the CFG residual as a critical and controllable failure channel. We proposed a retraining-free, sampler-side guidance recalibration method that adjusts only the CFG scale for a target CIM operating condition. Extensive experiments show that the preferred guidance scale increases with CIM noise and that our method consistently restores generation quality. At $\sigma_{\mathrm{CIM}}=0.20$, guidance recalibration closes at least $87\%$ of the CIM-induced FID gap, reducing FID from $59.22$ to $20.49$ on PixArt-$\Sigma$, from $72.37$ to $21.12$ on PixArt-$\alpha$, and from $20.89$ to $6.62$ on DiT-XL/2. We further provide a trajectory-level interpretation of how guidance recalibration restores generation quality under CIM noise.

\clearpage
\FloatBarrier
\bibliography{references}

\end{document}